\documentclass[conference]{IEEEtran}
\usepackage{times}

\usepackage[numbers]{natbib}
\usepackage{multicol}
\usepackage[bookmarks=true]{hyperref}
\usepackage{amssymb}
\usepackage{amsmath}
\usepackage{graphicx}

\usepackage{algorithm}
\usepackage{algorithmic}
\usepackage{booktabs}
\usepackage{multirow}
\usepackage{bigdelim}
\usepackage{subfigure}
\usepackage{url}

\DeclareMathOperator{\wrap}{wrap}
\DeclareMathOperator{\diag}{diag}

\begin{document}



\title{Occupancy-SLAM: Simultaneously Optimizing Robot Poses and Continuous Occupancy Map}

\author{\authorblockN{Liang Zhao, Yingyu Wang, and Shoudong Huang}
\authorblockA{Robotics Institute, Faculty of Engineering and Information Technology\\ University of Technology Sydney, Australia.\\ Email: Liang.Zhao@uts.edu.au}}



%

\maketitle

\begin{abstract}

In this paper, we propose an optimization based SLAM approach to simultaneously optimize the robot trajectory and the occupancy map using 2D laser scans (and odometry) information. The key novelty is that the robot poses and the occupancy map are optimized together, which is significantly different from existing occupancy mapping strategies where the robot poses need to be obtained first before the map can be estimated. In our formulation, the map is represented as a continuous occupancy map where each 2D point in the environment has a corresponding evidence value. The Occupancy-SLAM problem is formulated as an optimization problem where the variables include all the robot poses and the occupancy values at the selected discrete grid cell nodes. We propose a variation of Gauss-Newton method to solve this new formulated problem, obtaining the optimized occupancy map and robot trajectory together with their uncertainties. Our algorithm is an offline approach since it is based on batch optimization and the number of variables involved is large. Evaluations using simulations and publicly available practical 2D laser datasets demonstrate that the proposed approach can estimate the maps and robot trajectories more accurately than the state-of-the-art techniques, when a relatively accurate initial guess is provided to our algorithm. The video shows the convergence process of the proposed Occupancy-SLAM and comparison of results to Cartographer can be found at \url{https://youtu.be/4oLyVEUC4iY}.
\end{abstract}

\IEEEpeerreviewmaketitle

\section{Introduction}
Simultaneous localization and mapping (SLAM) is an important problem that involves building a map of an unknown environment and simultaneously estimating the location of the robot within the map \cite{cadena2016past}. With the recent development, the back-end of point feature-based SLAM \cite{dellaert2006square} 
is relatively mature, with both filter-based techniques and optimization-based techniques. In general, optimization-based techniques can result in the best estimate since the robot poses and the feature map are optimized together.

Occupancy maps are more useful for robot navigation, path planning and scene understanding since all the obstacles, free space and unknown areas in the environment are clearly marked in the map. 
When the robot poses used to collect the sensor information are known exactly, the evidence grid mapping technique \cite{elfes1989occupancy,moravec1985high,
moravec1989sensor} is an elegant and efficient approach to generate and update the occupancy grid map using the sensor data with the probabilistic model based on the Bayesian formulation. However, when robot navigates in an unknown environment and performs SLAM, its own poses need to be estimated and contain uncertainties. How to perform localization and build an occupancy map accurately at the same time is not trivial. 

In modern occupancy grid map based SLAM approaches such as Cartographer \cite{hess2016real}, the problem is solved in two steps. First, the robot poses are optimized using pose-graph SLAM, where the relative poses between the robot poses are constructed using scan matching, loop closure detection, or other similar techniques, and then a least squares optimization is performed to obtain an estimate of the robot trajectory. Second, the optimized poses are assumed to be the correct poses and are used to build up the map using evidence grid mapping techniques. In these two-step approaches, it is critical to obtain very accurate pose estimates in order to construct an accurate occupancy grid map. 
Similarly, other state-of-the-art occupancy mapping strategies such as Gaussian process (GP) based \cite{jadidi2018gaussian}, classification based \cite{ramos2016hilbert}, and Gaussian mixture model (GMM) based \cite{o2018variable}, all rely on a pre-estimated accurate robot trajectory. It can be expected that the occupancy map obtained using a two-step approach may not represent the best result that one can achieve using all the available information.

In this paper, we formulate the occupancy map based SLAM problem as an optimization problem where the robot poses and the occupancy map are optimized at the same time. A variation of Gauss-Newton method is proposed to solve the problem offline. Using our proposed Occupancy-SLAM method, more accurate robot poses and occupancy map can be obtained as compared with using existing occupancy mapping approaches, provided that a relatively accurate initial guess (e.g. the result obtained from Cartographer or GMapping\cite{grisetti2007improved}) is available. In addition, the corresponding uncertainty of the map can be obtained through the information matrix. Experimental results using simulation and practical datasets confirm the superior performance of the proposed SLAM method.


\section{Related Work}

Occupancy mapping has been studied for decades. Early works on simultaneous localization and occupancy grid mapping using laser scans 
are particle filter based. Particle filter based methods such as FastSLAM \cite{montemerlo2002fastslam} and GMapping have the advantage that the occupancy of each grid can be included in the state representation, since each particle includes a robot trajectory and an occupancy grid map (OGM) associated with it. However, due to the principle of particle filtering, particle filter based methods take up a lot of resources when the robot trajectory becomes long. Although some approaches try to address these shortcomings, such as GMapping proposes a novel resampling strategy, \citet{strom2011occupancy} suggest to compute submaps, these problems still exist. In general, filter-based approach can only achieve accurate results for short-term, but may fail for large-scale environments with long robot trajectories.

Optimization-based approaches, such as Hector SLAM \cite{kohlbrecher2011flexible} and Cartographer \cite{hess2016real}, aim to improve the accuracy of the poses by reducing the cumulative error of the SLAM system. Hector SLAM \cite{kohlbrecher2011flexible} avoids the accumulation of errors in scan matching by a scan-to-map matching. Specifically, it formulates a least squares problem by matching the endpoints of laser beams to submaps of the OGM and finds the locally optimal pose by a Gauss-Newton method. In addition, it achieves implicit matching with all previous scans by aligning the current scan with the existing map. However, Hector SLAM cannot reduce the accumulation of errors in the long term by loop closure detection, and it can only build short-term OGM but not large-scale OGM.

Based on the scan-to-map matching idea of Hector SLAM, Cartographer \cite{hess2016real} additionally supports loop closure detection to address the long-term global error accumulation. Specifically, all scans are first inserted into submaps using optimized poses by the scan-submap matching. When all submaps are constructed, submaps and scans are matched for detecting loop closures in the back-end, and the pose-graph including loop closure constraints is formulated as a least squares problem and solved. Other loop closure detection methods for grid mapping are also presented to eliminate the cumulative error, such as \cite{blanco2013robust}, where a map-to-map matching is used to detect the occurrence of loop closure to solve drift problem.

Besides OGM with fixed grid resolution, there are multiple ways to represent the occupancy map. For example, OctoMap \cite{hornung2013octomap} uses a memory efficient hierarchical structure to represent the grid map. Some other works represent the occupancy map as a continuous probability distribution model rather than a grid map to capture the structural information and the uncertainty of the environment which can be better used for tasks such as exploration. GMM is used in \cite{o2018variable} to represent the structure of the environments. MRFMap \cite{shankar2020mrfmap} uses a Markov Random Field model which utilises depth information to build the continuous 3D occupancy grid map in real-time. GP based method uses variance surface of Gaussian processes, corresponding to a continuous representation of uncertainty in the environment, to calculate the mean and variance of the occupancy map when the robot acquires a new observation, therefore it can be used to compute the continuous probability distribution in occupancy. The original formulation of GP based occupancy mapping, which is a batch method, was proposed by \cite{o2009contextual,o2012gaussian} and the incremental method was proposed by \cite{jadidi2018gaussian,kim2012building}. Hilbert Maps technique is proposed by \cite{ramos2016hilbert} for continuous occupancy mapping, in which the mapping problem is treated as a binary classification problem in a reproducing kernel Hilbert space.  Similarly, a more accurate classification based approach is proposed recently in \cite{liu2021efficient}.

All these non-filter based approaches need to optimize the robot poses first and then build the occupancy map based on the optimized poses, so the localization and mapping steps are not ``simultaneous". Although there are efficient and reliable pose-graph solvers (e.g. \cite{rosen2019se}) to optimize the robot poses, the global occupancy map information is not well considered when formulating the pose-graph problem, thus the results obtained using these two-step approaches are likely to be suboptimal. In this paper, we propose an optimization framework that can optimize the robot poses and the occupancy map ``simultaneously". Our occupancy map is continuous using the occupancy values at the selected grid cell nodes as parameters. After the optimization, the uncertainty of the robot poses and the occupancy map can also be obtained using the information matrix. 


Very recently, there are some research papers that  have tried to simultaneously optimize the poses and the map, but have used  different map representations. For example, Kimera-PGMO proposed in \cite{rosinol2021kimera} is a novel approach that simultaneously optimize the poses and the mesh representation of the environment. VoxGraph \cite{reijgwart2019voxgraph} represents the environment as a collection of overlapping signed distance function submaps, and optimizes the poses and the alignment of the submap collection simultaneously. These works have similar motivations as our paper, aiming to achieve better quality maps through joint optimization.

\section{Problem Formulation}\label{sec_3}

Our approach considers the simultaneous optimization of the robot poses as well as the occupancy map using 2D laser observations (and odometry). In this section we will explain how the observations from the laser can be linked to the robot poses and the occupancy map to formulate the nonlinear least squares (NLLS) problem. We also explain how we improve the problem formulation to make it easier to solve by adding a smoothing term.

\subsection{The Available Information}\label{Sec_Info}

The available information includes 2D laser scans collected at different robot poses. In addition, the odometry information might also be available. 

\subsubsection{Sampling Scan Points}\label{Sec_Info_1}

We first introduce our sampling strategy to generate the observations from the laser scans to be used in our NLLS formulation. 

\begin{figure}[t]
\centering \subfigure[Equidistant Sampling Strategy] {\label{fig_sampling_strategy}
\includegraphics[width=0.23\textwidth]{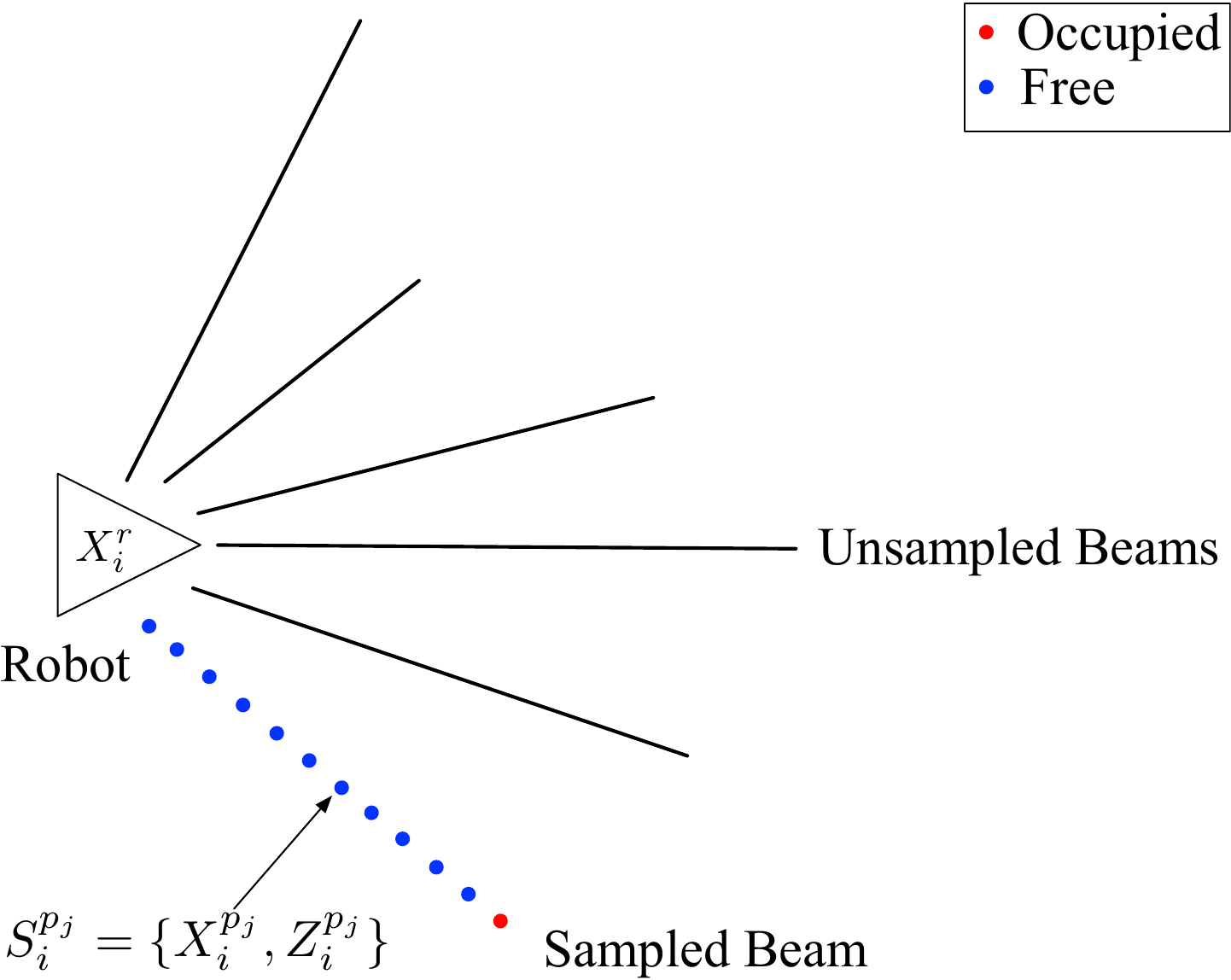}}
\centering \subfigure[Observation Points in one Scan] {\label{fig_scan}
 \includegraphics[width=0.23\textwidth]{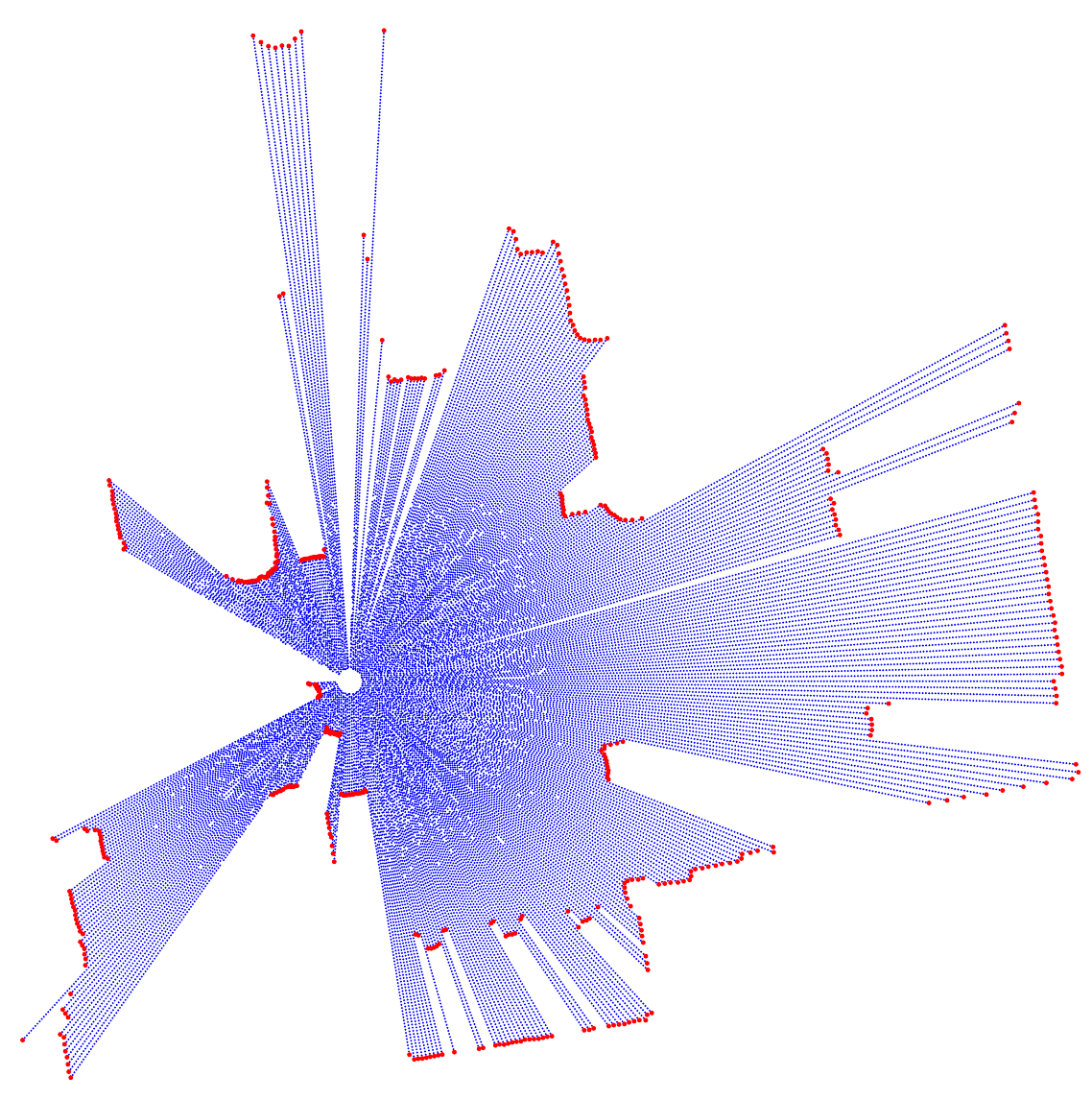}}
\caption{Sampling strategy for generating observations from a laser scan. (a) Observation points are sampled on a beam in the scan using equidistant sampling strategy. The red point indicates occupied state and the blue points indicate free states. The distance between two consecutive points is $s$ (the resolution). (b) All sampled observation points at a given time step.}
\label{fig_scan_sampling}
\end{figure}

Each scan data consists of a number of beams. On each beam, the endpoint  indicates the presence of an obstacle while the other points before the endpoint indicate the absence of obstacles. 
Here we sample each beam using a fixed resolution $s$ to get the observations, as shown in Fig. \ref{fig_sampling_strategy}.

Suppose there are $n+1$ laser scans in total collected from robot poses $0$ to $n$. By constant equidistant sampling of all the beams for the scan collected at timestep $i$, $(0 \leq i \leq n)$, a sampling point set
\begin{equation}
S_i=\{ S^{p_j}_i \triangleq \{X^{p_j}_i,Z_i^{p_j}\}\}~(1 \leq j \leq k_i)
\label{S_i}
\end{equation} 
can be obtained, where ${X}^{p_j}_i$ denotes the position of sampling point $p_j$ at time step $i$ in the local robot/laser coordinate frame and 
\begin{equation}
Z_i^{p_j} = \log \frac{p(S^{p_j}_i|{X}_i^{p_j} \in occ)}{1-p(S^{p_j}_i|{X}_i^{p_j} \in occ)}
\end{equation}
denotes the corresponding occupancy value. Here we use the evidence which is the log of odds (the ratio between the probability of being occupied and the probability of being free) \cite{martin1996robot}. In our implementation, we use $p(S^{p_j}_i|X_i^{p_j} \in occ) = 0.7$ for an occupied point (red in Fig. \ref{fig_sampling_strategy}) and $p(S^{p_j}_i|X_i^{p_j} \in occ) = 0.4$ for a free point (blue in Fig. \ref{fig_sampling_strategy}), resulting in $Z_i^{p_j} = 0.8473$ for an occupied point and $Z_i^{p_j} = -0.4055$ for an unoccupied point \cite{hornung2013octomap,ProbabilisticRobotics}. 
Fig. \ref{fig_scan} shows an example of the sampling point set for one scan.

$S_i$ $(0 \leq i \leq n)$ in (\ref{S_i}) will be used as observations in our NLLS formulation. It should be noted that since the total length of all the beams is different at different time step $i$, the number of sampling points $k_i$ obtained by the equidistant sampling strategy varies for different time step $i$. 

\subsubsection{Odometry Information}
The odometry information might be available for building the map as well. We assume the odometry input is the relative pose between two consecutive steps. 
The odometry from robot pose $X^r_{i-1}$ to pose $X^r_{i}$ can be represented as 
\begin{equation} 
O_i=\left[ (O_i^t)^T,O_i^\theta \right]^T~~(1 \leq i \leq n)
\label{O_i}
\end{equation}
where $O_i^t$ is the translation part and $O_i^\theta$ is the rotation angle part.  

\subsection{Continuous Occupancy Map $M(\cdot)$}
Similar to \cite{kohlbrecher2011flexible}, we regard the occupancy map $M$ as continuous. That is, the occupancy values are defined continuously on every point in the environment. This not only allows the derivatives to be calculated directly but also allow a more accurate representation of the position of the scan points when they are projected into the map using the robot poses. 

\begin{figure}
\centering
\includegraphics[width=0.4\textwidth]{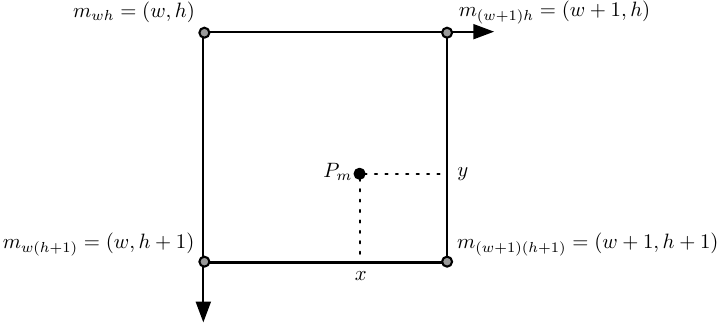}
\caption{\label{fig_interpolation} The bilinear interpolation method for a continuous coordinate $P_m$, whose occupancy value can be computed by those of the four adjacent grid cell nodes $m_{wh}, m_{({w+1})h}, m_{w({h+1})}, m_{({w+1})({h+1})}$.}
\end{figure}

Suppose the environment is divided into $l_w\times l_h$ grid cells. We use $m_{wh}=[w,h]^T~(0 \leq w \leq l_w, 0 \leq h \leq l_h)$ to represent the coordinate of a discrete cell node in the map and the occupancy value of this cell node is $M(m_{wh})$. Similar to \cite{gutmann2012vector}, the continuous occupancy map $M$ is defined uniquely using the $(l_w+1) \times (l_h+1)$  parameters $\{M(m_{wh})\}$ through bilinear interpolation of the  occupancy values at the four adjacent cell nodes, as shown in Fig. \ref{fig_interpolation}. 

Given an arbitrary coordinate $P_m=[x,y]^T$ in the occupancy map $M$ with four adjacent cell nodes $m_{wh}, \cdots, m_{(w+1)(h+1)}$, as shown in Fig. \ref{fig_interpolation}, the occupancy value $M(P_m)$ can be calculated as
\begin{equation}
	M(P_m)= \begin{bmatrix}
a_1b_1,a_0b_1,a_1b_0,a_0b_0
\end{bmatrix}\left[
\begin{aligned}
&M(m_{wh})\\&M(m_{(w+1)h})\\&M(m_{w(h+1)})\\&M(m_{(w+1)(h+1)})
\end{aligned}\right] \label{eq_1}
\end{equation}
in which 
\begin{equation}
\begin{aligned}
	a_0 &= x - w\\
	a_1 &= w+1 - x\\
	b_0 &= y - h\\
	b_1 &= h+1 - y .\\
\end{aligned} \label{eq_2}
\end{equation}


Using this map representation, estimating the continuous occupancy map $M$ is equivalent to estimating the $(l_w+1)\times (l_h+1)$ variables $\{M(m_{wh})\}$.

\subsection{The NLLS Formulation} 


In this paper, we use NLLS formulation to optimize the robot poses and the occupancy map simultaneously.

Suppose the $n+1$ robot poses are expressed by $\{X^r_i \triangleq [t_i^T,\theta_i]^T\} ~(0 \leq i \leq n)$ where $t_i$ is the x-y position of the robot and $\theta_i$ is the orientation with the corresponding rotation matrix $R_i=\begin{bmatrix}
\cos(\theta_i), \sin(\theta_i)\\ -\sin(\theta_i), \cos(\theta_i)
\end{bmatrix}$. As in most of the SLAM problem formulations, we assume the first robot pose defines the coordinate system, $X^r_0 \triangleq [0,0,0]^T$, so only the other $n$ robot poses are variables that need to be estimated. Thus in the NLLS formulation of the proposed Occupancy-SLAM problem, the state $\mathbf{X}$ contains the $n$ robot poses and the occupancy values of the discrete grid cell nodes $\{M(m_{wh})\}~(0 \leq w \leq l_w, 0 \leq h \leq l_h)$, i.e.
\begin{equation}\label{Eq_State}
\mathbf{X} \triangleq\left[ ({X}^r_1)^T, \cdots, (X^r_n)^T, M(m_{00}),\cdots,M(m_{l_wl_h}) \right]^T. \end{equation}

Then, the objective function of the NLLS optimization problem can be formulated as follows 
\begin{equation}
f(\mathbf{X})=w_Zf^Z(\mathbf{X})+w_Of^O(\mathbf{X})+w_S f^S(\mathbf{X}).	\label{eq_9}
\end{equation}
Here, the objective function consists of three parts: the observation term $f^Z(\mathbf{X})$, the odometry term $f^O(\mathbf{X})$ and the smoothing term $f^S(\mathbf{X})$. And $w_Z$, $w_O$ and $w_S$ are the weights for the different terms. The details are given in the following subsections.

\subsection{Observation Term}

Given the observation information $S = \{S_i\}$ $(0 
\leq i \leq n)$ in (\ref{S_i}), the observation term in the objective function (\ref{eq_9}) is formulated as
\begin{equation}
\begin{aligned}
	f^Z(\mathbf{X}) &=
	\sum_{i=0}^n \sum_{j=1}^{k_i}  \left\|Z_i^{p_j} - F_{ij}^Z(\mathbf{X})\right\|^2, ~\mbox{where} \\
	F_{ij}^Z(\mathbf{X})  &= \frac{M(P_{m_{i}^{p_j}})}{N({P_{m_{i}^{p_j}}})}.\\
\end{aligned}\label{obs-term}
\end{equation}
Here $P_{m_{i}^{p_j}}$ denotes a continuous coordinate in the occupancy map $M$ where the scan point $p_j$ is projected to the grid cells using the robot pose $i$, which can be calculated by
\begin{equation}
	P_{m_{i}^{p_j}}
=\frac{R_i^T X_{i}^{p_j}+t_i-t_0}{s} \label{P-project}
\end{equation}
where $s$ is the resolution of the cell nodes in the occupancy map $M$ (the distance between two adjacent cell nodes represents $s$ meters in the real world). Here we use the same resolution as used in generating observations from laser scans in Section \ref{Sec_Info_1}. And $t_0$ is the position of the occupancy map origin. $M(P_{m_{i}^{p_j}})$ is the occupancy value of point $P_{m_{i}^{p_j}}$ in the occupancy map $M$ and it can be calculated by (\ref{eq_1}). 

In (\ref{obs-term}), $N({P_{m_{i}^{p_j}}})$ is the equivalent hit multiplier of $P_{m_{i}^{p_j}}$ which will be described clearly in Section \ref{sec_hit}. Intuitively, since the evidence value is additive, the occupancy value $M(P_{m_{i}^{p_j}})$ is proportional to the number of times the point is observed.

Here, we suppose the errors of occupancy values of different sampled points in the observations $\{S_i\}$ are independent and have the same uncertainty. Therefore, the weights on all terms of the objective function (\ref{obs-term}) are the same, it is equivalent to set all the weights equal to $1$. Thus, we use norms instead of weighted norms in equation (\ref{obs-term}).


\subsection{Hit Map $N(\cdot)$}\label{sec_hit}
When using the Bayesian approach to build OGM, the occupancy of a cell is calculated as the sum of the occupancy values of all the scan points projected to this cell \cite{ProbabilisticRobotics}. Thus $N({P_{m_{i}^{p_j}}})$ is needed in (\ref{obs-term}) to present the relationship between the occupancy map $M$ and each individual observation $Z_i^{p_j}$. Now we explain how $N({P_{m_{i}^{p_j}}})$ can be calculated. 

The function $N(\cdot)$ is a hit map which indicates equivalently how many scan points are projected onto each continuous position in the map. It depends on the (current estimate of) robot poses.

When the robot poses $\{X^r_i\}~(0 \leq i \leq n)$ are given, all the scan points $\{X^{p_j}_i\}~(1 \leq j \leq k_i, 0 \leq i \leq n)$ in the observations $\{S_i\}$ can be projected into the grid cells using (\ref{P-project}) and each point contributes $1$ hit. Then, the equivalent hit multiplier at the grid cell nodes $\{N(m_{wh})\}$ can be computed by 
\begin{equation}
\left[ N(m_{00}),\cdots,N(m_{l_wl_h}) \right] 
= \sum_{i=0}^n \sum_{j=1}^{k_i} H(P_{m_{i}^{p_j}})
\label{eq_NP}
\end{equation}
where $P_{m_{i}^{p_j}}$ can be calculated using (\ref{P-project}), and $H(\cdot)$ is the function that distributes the $1$ hit to all the nodes, with only the four nearby nodes non-zero through the inverse of bilinear interpolation. 

After the hit map values at the cell nodes $\{N(m_{wh})\}$ are obtained, calculating the equivalent hit multiplier $N(P_{m_{i}^{p_j}})$ for an arbitrary continuous point (as used in the objective term in (\ref{obs-term})) can be easily obtained using the bilinear interpolation, which is similar as in (\ref{eq_1}).

\subsection{Odometry Term}
If odometry inputs $O = \{O_i\}~(1 \leq i \leq n)$ are available in the format of (\ref{O_i}), then the odometry term can be formulated as

\begin{equation}
\begin{aligned}
f^O(\mathbf{X})&=\sum_{i=1}^n \left\|O_i -
F_i^O(\mathbf{X})
\right\|^2_{\Sigma^{-1}_{O_i}}
\\&=\sum_{i=1}^n\left\|
\begin{bmatrix}
O_i^t-R_{i-1}\left(t_i - t_{i-1} \right)\\
\wrap\left(O_i^\theta- \theta_i + \theta_{i-1}\right)
\end{bmatrix}
\right\|^2_{\Sigma^{-1}_{O_i}}  \label{eq_7}
\end{aligned}
\end{equation}
in which $\Sigma_{O_i}$ is the covariance matrix representing the uncertainty of $O_i$, and $\wrap(\cdot)$ wraparounds the rotation angle to $[-\pi,\pi]$.

\subsection{Smoothing Term}

It can be easily found out that minimizing the objective function with only the observation term and the odometry term is not easy, since there are a large number of local minima. Especially, if the robot poses are far away from the global minimum, it is very difficult for an optimizer to converge to the correct solution. 

In order to enlarge the region of attraction and develop an algorithm that is robust to poor initial values, we introduce a smoothing term. This term makes the occupancy map $M$ more smooth and continuous for derivative calculation. The smoothing term requires the occupancy values of nearby cell nodes to be close to each other. In our case, based on the derivative calculation method we use, we penalize the difference between each cell node in the map $M$ and the two neighboring cell nodes to its right and below as
\begin{equation}
\begin{aligned}
f^S(\mathbf{X})
& =\left\|F^S(\mathbf{X}) \right\|^2\\
& = \sum_{w=0}^{l_w-1} \sum_{h=0}^{l_h-1}  \left\|\begin{bmatrix} M(m_{wh})-M(m_{{(w+1)}h})\\
M(m_{wh})-M(m_{{w}{(h+1)}})
\end{bmatrix} \right\|^2 \\
& + \sum_{w=l_w} \sum_{h=0}^{l_h-1}  \left\| M(m_{wh})-M(m_{{w}{(h+1)}})\right\|^2 \\
& + \sum_{w=0}^{l_w-1} \sum_{h=l_h}  \left\| M(m_{wh})-M(m_{{(w+1)}{h}})\right\|^2.
\end{aligned} \label{eq_8}
\end{equation}
It should be noted that $F^S(\textbf{X})$ is a linear function of $\{M(m_{wh})\}$ in the state. The coefficient matrix is constant and can be calculated prior to the optimization. For more details, please refer to Section \ref{Sec_J_S}.

In our Occupancy-SLAM algorithm, we start from a larger weight for the smoothing term such that the robot poses can converge to near correct poses and then gradually reduce the weight such that the algorithm can converge to a more accurate map (the real occupancy map is not smooth). This strategy is similar to the graduated non-convexity idea used in \cite{yang2020graduated}.

\section{Proposed Occupancy-SLAM Algorithm}
In Section \ref{sec_3}, we introduced our NLLS formulation for the Occupancy-SLAM problem. In this section, we provide the details of the solution technique of Occupancy-SLAM, including the calculation of Jacobians which are different from traditional SLAM. 


\subsection{Iterative Solution to the NLLS Formulation}



We denote the state (\ref{Eq_State}) as $\mathbf{X} = \left[(\mathbf{X}^r)^T,\mathbf{M} \right]^T$, where
\begin{equation}
\begin{aligned}
\mathbf{X}^r &= \left[ (X^r_1)^T, \cdots, (X^r_n)^T \right]^T\\
\mathbf{M} &= \left[M(m_{00}),\cdots,M(m_{l_wl_h}) \right]^T.\\
\end{aligned}
\end{equation}
Here $\mathbf{X}^r$ contains all the robot poses and $\mathbf{M}$ contains all the occupancy values at the grid cell nodes. Let
\begin{equation}
\begin{aligned}
F(\mathbf{X}) = [&\cdots,Z_i^{p_j}-F_{ij}^Z(\mathbf{X}),\cdots,(O_i-F_i^O(\mathbf{X}))^T,\\
&\cdots,F^S(\mathbf{X})^T]^T\\
W = \;\; &\diag(\cdots,w_Z,\cdots,w_O \Sigma^{-1}_{O_i}, \cdots,w_S, \cdots)\\
\end{aligned}
\end{equation}
combine all the error functions and the weights of the three terms in (\ref{eq_9}). Then, the NLLS optimization problem in (\ref{eq_9}) is to seek $\mathbf{X}$ such that
\begin{equation}\label{Least Squares}
f(\mathbf{X})=\|F(\mathbf{X})\|^2_W =
F(\mathbf{X})^T W
F(\mathbf{X})
\end{equation}
is minimized.

A solution to (\ref{Least Squares}) can be obtained iteratively by starting with an initial guess $\mathbf{X}(0)$ and updating with $\mathbf{X}(k+1) = \mathbf{X}(k) + \Delta(k)$. The update vector $\Delta (k) = [\Delta^r(k)^T,\Delta^M(k)]^T$ is the solution to
\begin{equation}\label{Gauss-Newton}
J^T W J \Delta (k) = -J^T W F(\mathbf{X}(k))
\end{equation}
where $J$ is the linear mapping represented by the Jacobian matrix
$\partial F / \partial \mathbf{X}$ evaluated at $\mathbf{X}(k)$.

After the optimal solution $\hat{\mathbf{X}}$ of the robot poses and the occupancy map is obtained, the corresponding covariance matrix $\Sigma$ can be computed by $\Sigma^{-1} = J^T W J$ where the Jacobian $J$ is evaluated at $\hat{\mathbf{X}}$.

When solving the optimization problem, initially the smoothing term is given by a relative larger weight $w_S$ in order to make the algorithm more robust to the errors of initial guess. And this weight will be further reduced during the iterations ($w_S$ is reduced by $d_S$ times after $\tau_S$ iterations) to obtain more accurate occupancy map results. The iterative solution to the proposed Occupancy-SLAM is shown in Algorithm \ref{alg_1}, in which $\tau_k$ and $\tau_{\Delta}$ represent the thresholds of iteration number $k$ and the incremental vector $\Delta$.


\begin{algorithm}[h]  
  \caption{Our Occupancy-SLAM Algorithm}  
  \label{alg_1}  
  \begin{algorithmic}[1]
    \REQUIRE  
      Observations $S$, 
      Odomotry $O$,  
      Initial poses $X^r(0)$
    \ENSURE  
     Optimal poses $\hat{X}^r$ and map $\hat{M}$, covariance $\Sigma$
    \STATE Initialize occupancy of cell nodes $\mathbf{M}(0)$, hit map $N(0)$ using $X^r(0)$ and $S$
    \STATE Pre-calculate smoothing term coefficient matrix $A$
    \FOR {$k=1$; $k <= \tau_k \; \& \; \| \Delta(k) \|^2 >= \tau_{\Delta}$; $k++$}
      \IF {$k~\%~\tau_S=0$} 
        \STATE $w_S = w_S/d_S$
      \ENDIF
      \STATE Calculate gradient $\nabla M(k)$ of $\mathbf{M}(k)$
      \STATE Evaluate $F(\mathbf{X})$ and Jacobian $J$ at $X^r(k)$ and $\mathbf{M}(k)$
      \STATE Solve $J^T W J \Delta(k) =-J^T W F(\mathbf{X})$, where $\Delta(k) = [\Delta^r(k)^T,\Delta^M(k)^T]^T$
      \STATE Update poses $X^r(k+1)=X^r(k) + \Delta^r(k)$ and occupancy of cell nodes $\mathbf{M}(k+1)=\mathbf{M}(k)+\Delta^M(k)$
      \STATE Recalculate hit map $N(k+1)$ using $X^r(k+1)$ and $S$
      
    \ENDFOR  
    \STATE  $\hat{X}^r = X^r(k)$, $\hat{M} \Leftarrow \mathbf{M}(k)$, $\Sigma^{-1} = J^T W J$
  \end{algorithmic}  
\end{algorithm}

As both the robot poses and the occupancy map are optimized simultaneously, the Jacobian $J$ in (\ref{Gauss-Newton}) is very important and quite different from those used in the traditional SLAM algorithms. The Jacobian $J$ consists of four parts, i.e. the Jacobian of the observation term w.r.t. the robot poses $J_P$, the Jacobian of the observation term w.r.t. the occupancy map $J_M$, the Jacobian of the odometry term w.r.t. robot poses $J_O$ and the Jacobian of the smoothing term w.r.t. the occupancy map $J_S$. In the following subsections, we will detail how the Jacobians can be analytically calculated, which are used in the proposed Occupancy-SLAM algorithm.

\subsection{Jacobian of the Observation Term w.r.t. Robot Poses}

The Jacobian $J_P$ of function $F_{ij}^Z(\mathbf{X})$ in the observation term w.r.t. the robot poses $X^r$ can be calculated by the chain rule
\begin{equation}
	\begin{aligned}
		J_P=\frac{ \partial F_{ij}^Z(\mathbf{X}) }{ \partial X_i^r } = \frac{\partial F_{ij}^Z(\mathbf{X}) }{ \partial P_{m_i^{P_j}} } \cdot \frac{\partial P_{m_i^{P_j}}  }{ \partial X_i^r}	
	\end{aligned}
\end{equation}
in which $\dfrac{\partial P_{m_i^{P_j}}  }{ \partial X_i^r}$ can be calculated as
\begin{equation}
\dfrac{\partial P_{m_i^{P_j}}}{\partial X_i^{r}}=\left[\begin{array}{ll}
\dfrac{\partial P_{m_i^{P_j}}}{\partial t_i} & \dfrac{\partial P_{m_i^{P_j}}}{\partial \theta_i}
\end{array}\right]=\dfrac{1}{s} \left[\begin{array}{ll}
E_{2} & \left(R_i^{\prime}\right)^{T} X^{p_j}_i
\end{array}\right].
\end{equation}
$R_i^\prime$ is the derivative of the rotation matrix $R_i$ w.r.t. rotation angle $\theta_i$ and $E_2$ means $2 \times 2$ identity matrix.


$\dfrac{\partial F_{ij}^Z(\mathbf{X}) }{ \partial P_{m_i^{P_j}} }$ can be calculated by
\begin{equation}
\dfrac{\partial F_{ij}^Z(\mathbf{X}) }{ \partial P_{m_i^{P_j}} } = \dfrac{1}{N(P_{m_i^{P_j}})} \dfrac{\partial M(P_{m_i^{P_j}})}{\partial P_{m_i^{P_j}}}.
\end{equation}
Here $\dfrac{\partial M(P_{m_i^{P_j}})}{\partial P_{m_i^{P_j}}}$ can be considered as the gradient of the occupancy map $M$ at point $P_{m_i^{P_j}}$, which can be approximated by the bilinear interpolation of the gradients of the occupancy at the four adjacent cells $\nabla M(m_{wh}),\cdots,\nabla M(m_{(w+1)(h+1)})$ around $P_{m_i^{P_j}}$ as
\begin{equation}
\dfrac{\partial M(P_{m_i^{P_j}})}{\partial P_{m_i^{P_j}}}= 
\left[
\begin{aligned}
a_1b_1\\a_0b_1\\a_1b_0\\a_0b_0\\
\end{aligned}\right]^T
\left[
\begin{aligned}
&\nabla M(m_{wh})\\&\nabla M(m_{(w+1)h})\\&\nabla M(m_{w(h+1)})\\&\nabla M(m_{(w+1)(h+1)})
\end{aligned}\right]\label{eq_14}
\end{equation} 
where the gradient of occupancy map $M$ at all the grid cell nodes $\nabla M$ can be easily calculated from $\{M(m_{wh})\} (0 \leq w \leq l_w, 0 \leq h \leq l_h)$ in the state. The bilinear interpolation used in (\ref{eq_14}) is similar to the method in (\ref{eq_1}) and (\ref{eq_2}).

Here we assume the robot poses $\{X_i^r\}$ change slightly in each iteration, so as to reduce the computational complexity, the hit map $N$ is considered as constant and updated using the current robot poses in each iteration. Thus, the derivative of $N(P_{m_i^{P_j}})$ is not calculated. 



\subsection{Jacobian of the Observation Term w.r.t. Occupancy Map}
Based on (\ref{eq_1}), the Jacobian $J_M$ of function $F_{ij}^Z(\mathbf{X})$ in the observation term w.r.t. the cell nodes of the occupancy map $\mathbf{M}$ can be calculated as
\begin{equation}
\begin{aligned}
J_M & = \dfrac{\partial F_{ij}^Z(\mathbf{X})}{\partial \left[ M(m_{wh}),\cdots, M(m_{(w+1)(h+1)}) \right]^T}\\
&= \dfrac{1}{N(P_{m_i^{P_j}})}\dfrac{\partial M(P_{m_i^{P_j}})}{\partial \left[ M(m_{wh}),\cdots, M(m_{(w+1)(h+1)}) \right]^T}\\ 
&= \dfrac{\begin{bmatrix}
a_1b_1,a_0b_1,a_1b_0,a_0b_0
\end{bmatrix}}{N(P_{m_i^{P_j}})}
\end{aligned}
\end{equation}
where $M(m_{wh}), \cdots, M(m_{(w+1)(h+1)})$ are the four nearest  cell nodes to $P_{m_i^{P_j}}$, and $a_0,a_1,b_0$ and $b_1$ are defined in (\ref{eq_2}).


\subsection{Jacobian of the Odometry Term}
The Jacobian $J_O$ of function $F_i^O(\mathbf{X})$ in the odometry term (\ref{eq_7}) is the partial derivative w.r.t. the robot poses $X^r$ since it is not related to the occupancy map in the state vector $\mathbf{X}$. Therefore, the Jacobian $J_O$ can be calculated as
\begin{equation}
\begin{aligned}
J_O &= \frac{\partial F_i^O(\mathbf{X})}{\partial \left[ (X_{i-1}^r)^T, (X_i^r)^T \right]^T }\\ 
&=\begin{bmatrix}
	 \dfrac{\partial F_i^O(\mathbf{X})}{\partial t_{i-1}} &
	 \dfrac{\partial F_i^O(\mathbf{X})}{\partial \theta_{i-1}} &
	 \dfrac{\partial F_i^O(\mathbf{X})}{\partial t_i} &
	 \dfrac{\partial F_i^O(\mathbf{X})}{\partial \theta_i}
 \end{bmatrix} 
 \\
 &=\begin{bmatrix}
 	-R_{i-1} & R_{i-1}^\prime(X^r_i-X^r_{i-1}) & R_{i-1} &\mathbf{0}_2\\
 	\mathbf{0}_2^T & -1 & \mathbf{0}_2^T & 1\\
 \end{bmatrix}
\end{aligned}
 \end{equation}
in which $\mathbf{0}_2$ means $2 \times 1$ zero vector.
 
\subsection{Jacobian of Smoothing Term}\label{Sec_J_S}

The Jacobian $J_S$ of function $F^S(\mathbf{X})$ in the smoothing term is the derivative of (\ref{eq_8}) w.r.t. the occupancy values at the cell nodes 
$\{M(m_{wh})\} ~(0 \leq w \leq l_w, 0 \leq h \leq l_h)$ 
due to it is not related to the robot poses in the state vector $\mathbf{X}$. It should be mentioned that $F^S(\mathbf{X})$ is linear w.r.t. $\{M(m_{wh})\}$
\begin{equation}
F^S(\mathbf{X}) = A \left[ M(m_{00}),\cdots,M(m_{l_wl_h}) \right]^T
\end{equation}
where the $(2l_wl_h+l_w+l_h) \times ((l_w+1)(l_h+1))$ coefficient matrix $A$ is sparse and with nonzero elements $1$ or $-1$. An example of the coefficient matrix can be shown as
\begin{equation}
	A = \begin{bmatrix}
    \vdots &\vdots  &\vdots  &\vdots  &\vdots  &\vdots  &\vdots &\vdots\\
 	\mathbf{0}^T & 1 & -1 & 0 & \mathbf{0}^T & 0 & 0 & \mathbf{0}^T\\
 	\mathbf{0}^T & 1 & 0  & 0 & \mathbf{0}^T & -1 & 0 & \mathbf{0}^T\\
 	\mathbf{0}^T & 0 & 1 & -1 & \mathbf{0}^T & 0 & 0 & \mathbf{0}^T\\
 	\mathbf{0}^T & 0 & 1 & 0 & \mathbf{0}^T & 0 & -1 & \mathbf{0}^T\\
    \vdots &\vdots  &\vdots  &\vdots  &\vdots  &\vdots  &\vdots &\vdots\\
 \end{bmatrix}.
\end{equation}
Here $\mathbf{0}$ represents a zero vector with appropriate dimensions. Therefore, the Jacobian of the smoothing term can be calculated as
\begin{equation}
J_S = \frac{\partial F^S(\mathbf{X})}{\partial \mathbf{M} } = A.\\ 
\end{equation}
Since $A$ is constant, $J_S$ can be pre-calculated and directly used in the optimization as shown in Algorithm \ref{alg_1}.
  
\section{Experimental Results}

In this section, we present the results of our algorithm using a number of datasets. First, due to the lack of ground truth in the practical datasets, we evaluate our algorithm qualitatively and quantitatively through simulation experiments and compare it with Cartographer \cite{hess2016real}. Secondly, we compare the results from our approach with those from Cartographer qualitatively using a number of practical datasets. 
Thirdly, we visually compare our results with some other recent occupancy mapping strategies using the Intel dataset. Finally, we demonstrate the robustness of our algorithm to poor initializations. Due to the different sensor conﬁgurations of different datasets, we adjust some parameters in Cartographer to achieve its best performance possible. In our algorithm, we choose $w_Z=1$, $w_O=1$, initial $w_S = 0.1$, $\tau_S=18$, and $d_s=10$. 
        
Our algorithm is currently implemented in MATLAB. Depending on the datasets, it takes 20-60 iterations for our algorithm  to converge and each iteration takes 20 seconds to a few minutes. Please note that we are not aiming for a real-time online SLAM, but an offline SLAM to achieve the best results possible. 

\subsection{Simulation Experiments}\label{simu_experiment}


\begin{figure}
\centering
\includegraphics[width=0.48\textwidth]{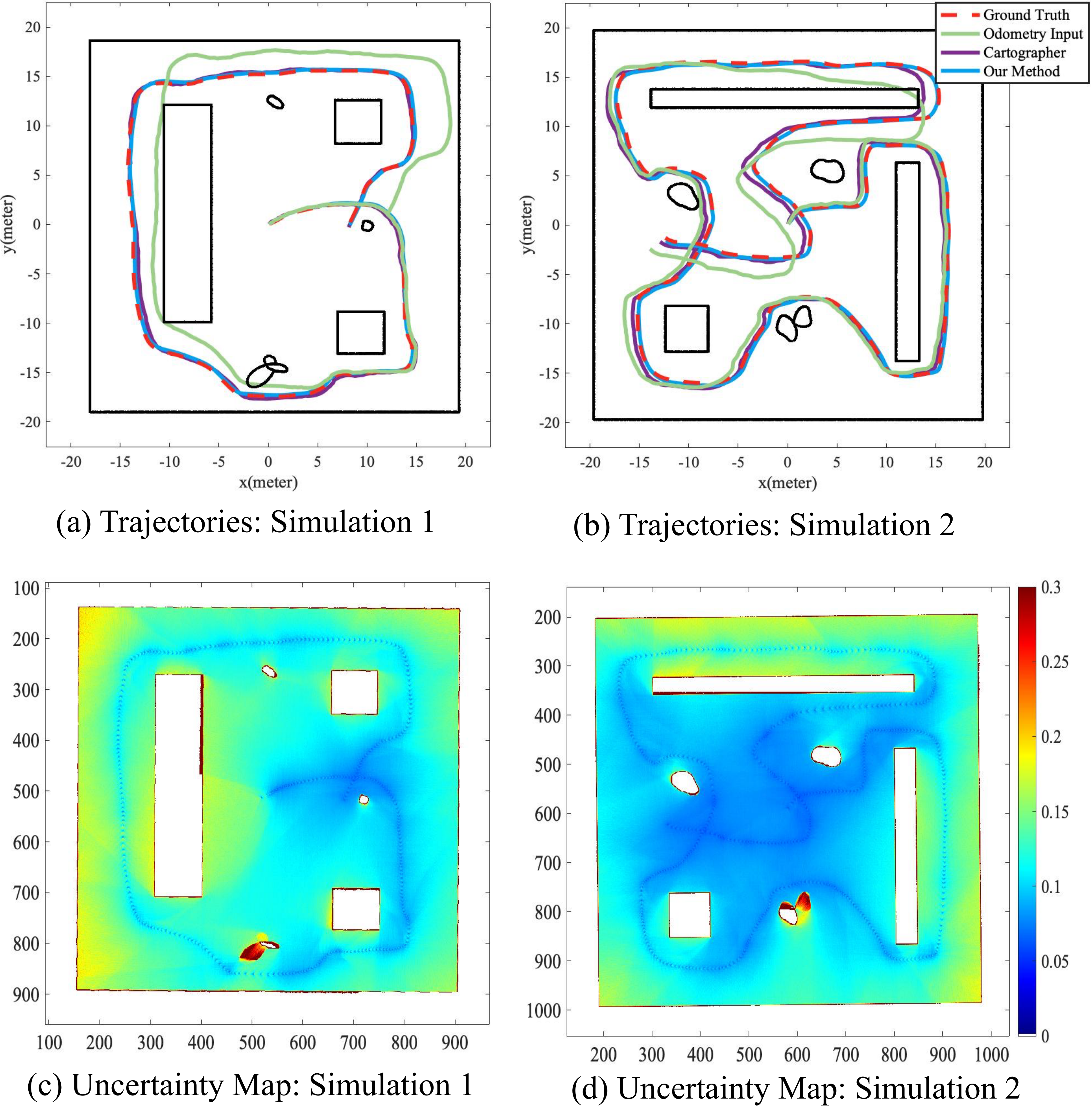}
\caption{\label{fig_trajectory_compare} Simulation environments, robot trajectory results and uncertainty maps. (a) and (b) show the simulation environments (the black lines indicate the obstacles in the scene) and the trajectories of ground truth, odometry inputs, Cartographer and our approach for one dataset in each of the two simulation experiments. (c) and (d) show the uncertainty maps of our approach for the same datasets. }
\end{figure}

We use two different simulation environments to design two different simulation experiments. Fig. \ref{fig_trajectory_compare}(a) and Fig. \ref{fig_trajectory_compare}(b) show the obstacles in the environments and the robot trajectories for the two experiments. Simulation 1 includes 340 scans collected from 340 robot poses, and with 339 odometry inputs (0.5 second per step). Simulation 2 includes 527 scans and 526 odometry inputs. Each scan consists of 1081 laser beams with angle ranging from -135 degrees to 135 degrees which simulates a Hokuyo UTM-30LX laser scanner. To simulate real-world data acquisition, we added random Gaussian noises with zero-mean and standard deviation of $0.02 m$ to each beam of the scan data generated from the ground truth. Similarly, we added zero-mean Gaussian noises to the odometry inputs generated from the ground truth poses (standard deviation of $0.04 m$ for x-y and $0.003 rad$ for orientation). For each simulation experiment, we generated 5 datasets each with different sets of random noises. In these experiments, our method uses the poses from Cartographer's results for initialization. 

\begin{figure}[t]
\centering \subfigure[Simulation 1] {\label{fig_time_error_1}
\includegraphics[width=0.48\textwidth]{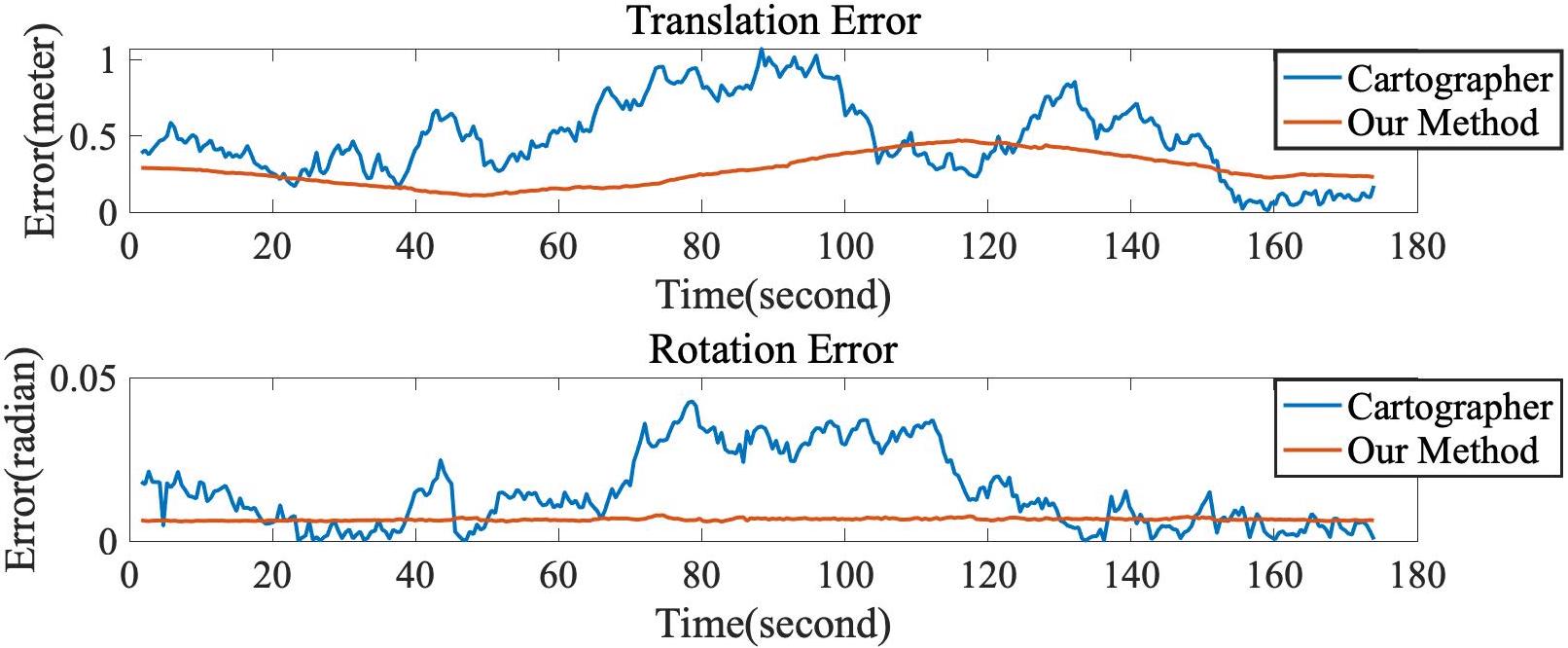}}
\centering \subfigure[Simulation 2] {\label{fig_time_error_2}
\includegraphics[width=0.48\textwidth]{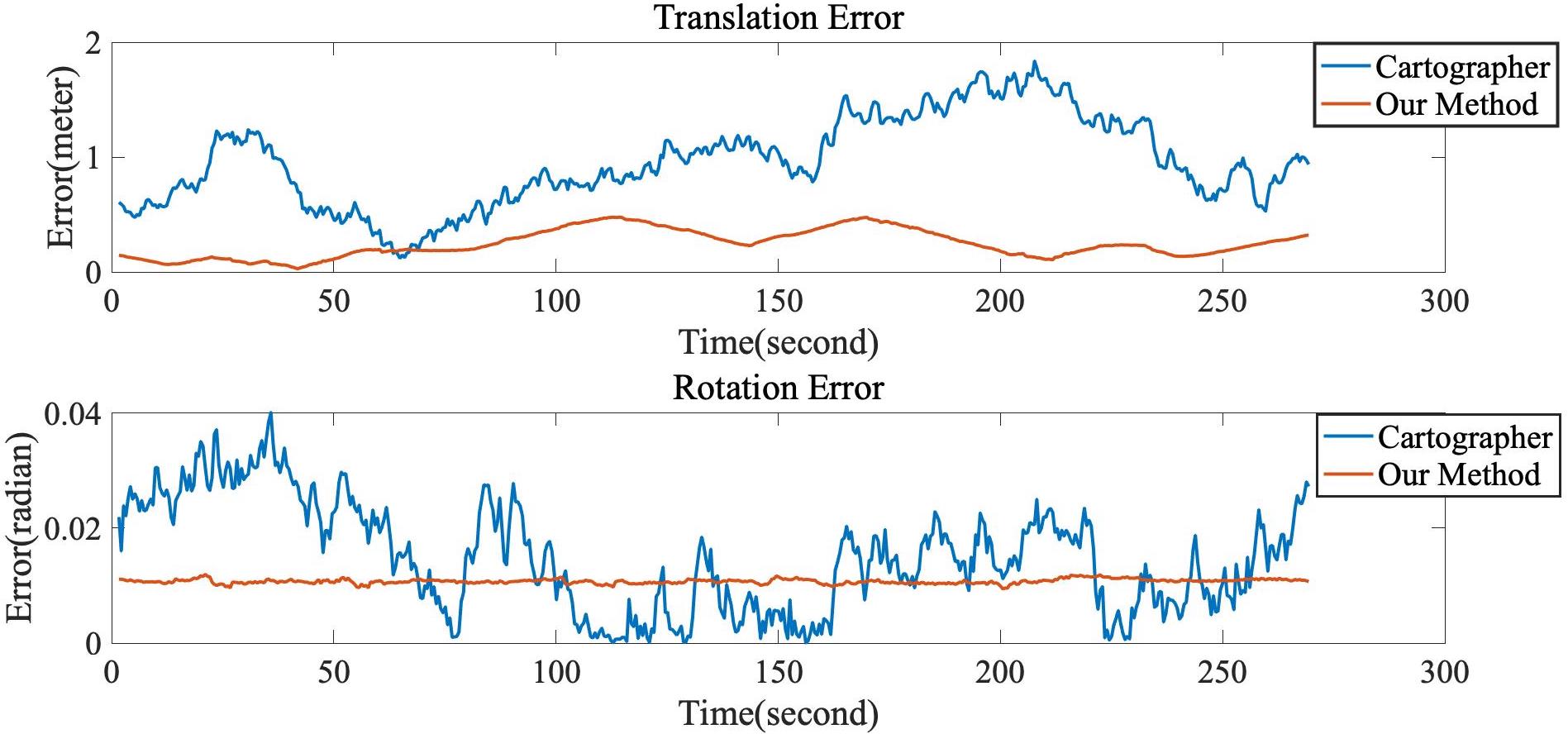}}
\caption{Comparison of Translation Error and Rotation Error.}
\label{fig_error_compare_time}
\end{figure}

The robot trajectory results of our Occupancy-SLAM and Cartographer of one dataset in each simulation are compared with the ground truth and odometry in Fig. \ref{fig_trajectory_compare}(a) and Fig. \ref{fig_trajectory_compare}(b). It can be seen that our trajectories are closer to the ground truth trajectories, especially for positions where significant rotation occurs. Fig. \ref{fig_error_compare_time} shows the translation and rotation errors of our method and Cartographer. It is clear that the errors of our method are smaller than those of Cartographer for the majority of the time.

We use all the five datasets for Simulation 1 and all the five datasets for Simulation 2 to perform quantitative comparison of pose estimation errors. The results are given in Table \ref{tab_comparison}. We use mean absolute error (MAE) and root mean squared error (RMSE) to evaluate the translation errors (in meters) and rotation errors (in radians). Our method performs best in all the four metrics for both simulations. 

\begin{table}[t]
		\centering
		\caption{Quantitative Comparison of Robot Pose Errors in Simulations}
		\label{tab_comparison}
		\setlength{\tabcolsep}{2mm}{
		\begin{tabular}{lccc}\toprule
			& Odometry Input & Cartographer & Our Method \\ \hline
		Simulation 1& & &\\
		\quad MAE of Translation & 1.5132 & 0.2428 & \textbf{0.1374}\\
		\quad MAE of Rotation & 0.0545 & 0.0149 & \textbf{0.0066}\\
		\quad RMSE of Translation & 1.9904 & 0.2871 & \textbf{0.1540}\\
		\quad RMSE of Rotation & 0.0657 & 0.0189 & \textbf{0.0066}\\\hline
		
		Simulation 2& & &\\
		\quad MAE of Translation & 0.8748 & 0.4779 & \textbf{0.1215}\\
		\quad MAE of Rotation & 0.0418 & 0.0148 & \textbf{0.0107}\\
		\quad RMSE of Translation & 1.0644 & 0.5542 & \textbf{0.1452}\\
		\quad RMSE of Rotation & 0.0514 & 0.0175 & \textbf{0.0107}\\\hline
		\end{tabular}
		}
\end{table}

\begin{figure*}
\centering
\includegraphics[width=1\textwidth]{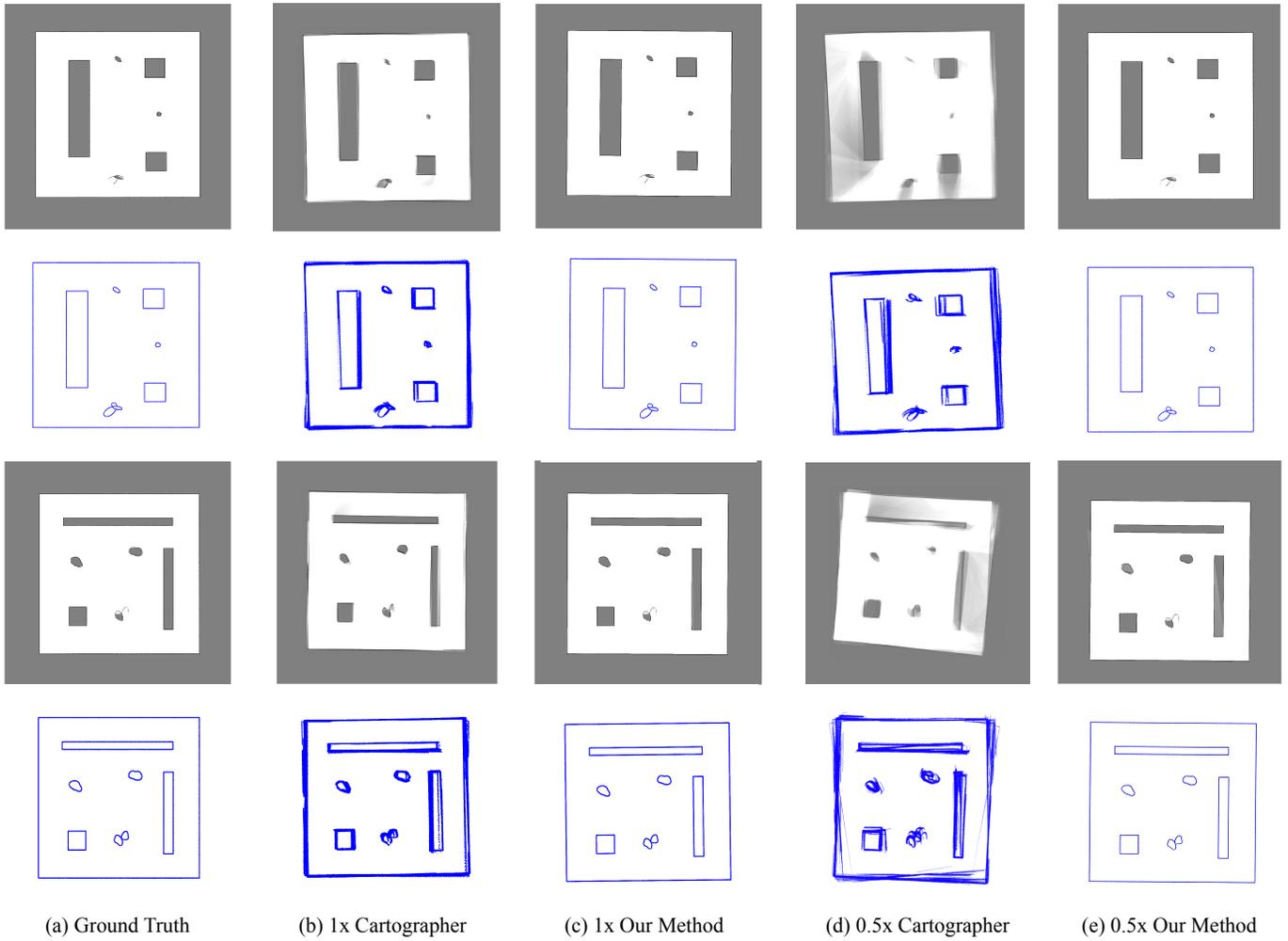}
\caption{\label{fig_simulation} The OGMs and point cloud maps generated by ground truth poses, poses from Cartographer and poses from our approach in one dataset for each simulation. The point cloud maps are generated by projection of the scan endpoints using the poses. The first two rows are the OGMs and point cloud maps of the dataset in Simulation 1, and the third to fourth rows are the OGMs and point cloud maps of the dataset in Simulation 2.
(b) and (c) show the results of original sampling rate. (d) and (e) show the results of 0.5 times sampling rate.}
\end{figure*}

Fig. \ref{fig_simulation} shows the OGMs and point cloud maps generated using poses from ground truth, Cartographer and our approach for one dataset from each of the simulations. It can be seen that the results of our method are far clearer than Cartographer's in terms of both OGMs and point cloud maps boundaries, which indicates that our method can obtain more accurate results by optimizing the robot poses and the occupancy map together. The uncertainty maps obtained by our method are given in Fig. \ref{fig_trajectory_compare}(c) and Fig. \ref{fig_trajectory_compare}(d).

\begin{figure}[h]
\centering
\includegraphics[width=0.48\textwidth]{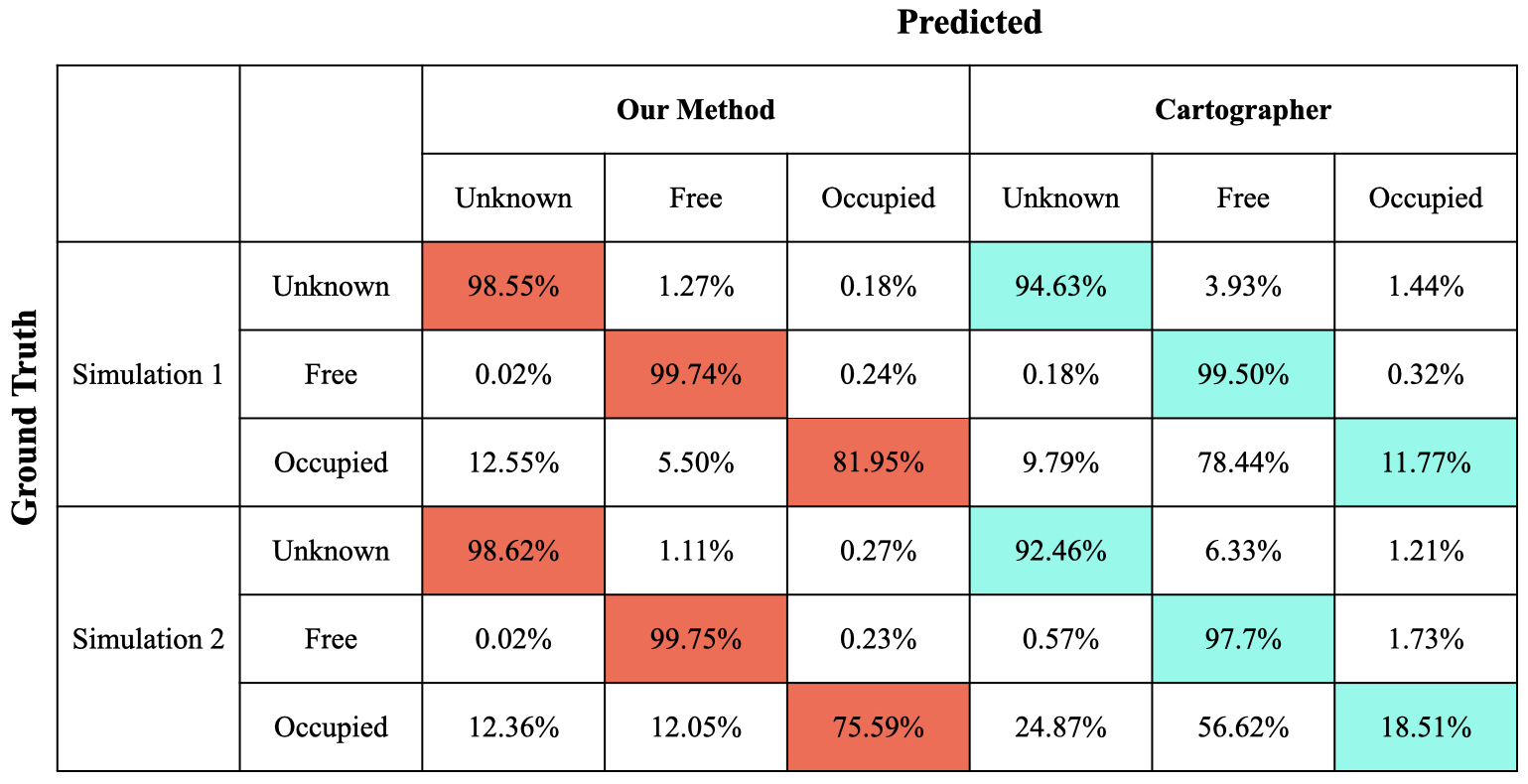}
\caption{\label{fig_predicted} The OGM precision of our method and Cartographer. }
\end{figure}

For quantitative comparison of the occupancy maps, we use the results from all the ten datasets in Simulation 1 and 2 to compare the accuracy of the maps. We regard the mapping problem as a classification problem where each grid cell is classified as either Free, Occupied or Unknown. The mapping performance of our method and Cartographer is given in   Fig. \ref{fig_predicted}. It should be noted that the low precision in classifying the Occupied cells does not necessarily mean the mapping quality is poor, since only the boundaries of the obstacles are Occupied cells.

We also use AUC (Area under the ROC curve) and precision to evaluate our method and Cartographer, in which labels of ground truth are given by occupancy map generated by ground truth poses. The results are given in Table \ref{tab_auc}. It can be seen that our method achieves better performance in both the metrics. For this evaluation method, we have removed all the unknown cells since AUC is a metric for binary classification.

\begin{table}[h]
		\centering
		\caption{Accuracy of the Occupancy Map}
		\label{tab_auc}
		\setlength{\tabcolsep}{2mm}{
		\begin{tabular}{lcccc}\toprule
		&\multicolumn{2}{c}{Simulation 1} &\multicolumn{2}{c}{Simulation 2}\\ 
		& Cartographer & Our Method & Cartographer & Our Method \\ \hline
		AUC & 0.9616  & \textbf{0.9968} & 0.8663  & \textbf{0.9881}\\
		Precision & 0.9619 & \textbf{0.9896} & 0.9415 & \textbf{0.9892}\\ \hline
		\end{tabular}
		}
\end{table}

%


Comparing to Cartographer, another advantage of our method is that higher sampling rates of laser scans and odometry inputs are not required for our approach. We reduced the sampling frequency of the observations and odometry inputs to 0.5 times of the original inputs and we use the poses from Cartographer for initialization in this experiment. It is shown in Fig. \ref{fig_simulation}(d) that the quality of Cartographer's results decreases significantly as the sampling rate decreases, which is caused by the low sampling frequency of the observations that does not allow Cartographer to generate reliable matching between scans and map. In the low sampling dataset for Simulation 2, due to the significant rotation of the robot in the early stages of motion, it caused Cartographer to produce significant mis-matching, which resulted in an overall shift in the generated OGM. It is also shown in Table \ref{tab_low_sampling}, the translation-related errors given by the Cartographer are significantly higher than the odometry input due to the shifts occur in the early stages, and our method reduces these errors significantly. 

\begin{table}[h]
		\centering
		\caption{Quantitative Comparison of Robot Pose Errors in the Low Sampling Rate Experiments}
		\label{tab_low_sampling}
		\setlength{\tabcolsep}{2mm}{
		\begin{tabular}{lccc}\toprule
			& Odometry Input & Cartographer & Our Method \\ \hline
		Simulation 1& & &\\
		\quad MAE of Translation & 1.6051 & 0.4498 & \textbf{0.1764}\\
		\quad MAE of Rotation & 0.0512 & 0.0383 & \textbf{0.0147}\\
		\quad RMSE of Translation & 2.0684 & 0.5250 & \textbf{0.2087}\\
		\quad RMSE of Rotation & 0.0632 & 0.0402 & \textbf{0.0147}\\\hline
		
		Simulation 2& & &\\
		\quad MAE of Translation & 0.8745 & 1.3636 & \textbf{0.7758}\\
		\quad MAE of Rotation & 0.0419 & 0.0308 & \textbf{0.0086}\\
		\quad RMSE of Translation & 1.0640 & 1.4481 & \textbf{0.7840}\\
		\quad RMSE of Rotation & 0.0515 & 0.0404 & \textbf{0.0114}\\\hline
		\end{tabular}
		}
\end{table}

\subsection{Ablation Study}
In Section \ref{simu_experiment}, our approach is shown to be effective through experiments on two simulation datasets. In this section, ablation experiments are conducted to further understand the benefits of jointly  optimizing the poses and grid map and the benefit of using continuous map representation. To this end, we perform experiments that jointly optimizing the poses and the discrete map. All the parameters are set the same as those used in Section \ref{simu_experiment}.

\begin{table}[h]
		\centering
		\caption{Ablation Study}
		\label{Ablation Study}
		\setlength{\tabcolsep}{2mm}{
		\begin{tabular}{lccc}\toprule
			 &Cartographer & Discrete Map & Full Method \\ \hline
		Simulation 1& & &\\
		\quad MAE of Translation  &0.2428& 0.1377 & \textbf{0.1374}\\
		\quad MAE of Rotation  &0.0149& 0.0117 & \textbf{0.0066}\\
		\quad RMSE of Translation  &0.2871& 0.1591 & \textbf{0.1540}\\
		\quad RMSE of Rotation  &0.0189& 0.0118 & \textbf{0.0066}\\\hline
		
		Simulation 2& & &\\
		\quad MAE of Translation & 0.4779& 0.1785 & \textbf{0.1215}\\
		\quad MAE of Rotation  &0.0148& 0.0134 & \textbf{0.0107}\\
		\quad RMSE of Translation  &0.5542& 0.1849 & \textbf{0.1452}\\
		\quad RMSE of Rotation  &0.0175& 0.0134 & \textbf{0.0107}\\\hline
		\end{tabular}
		}
\end{table}

The results are shown in Table \ref{Ablation Study}. For the discrete map representation method, the continuous map representation in our full Occupancy-SLAM approach is replaced by the discrete map representation but the joint NLLS optimization is reserved. The results show that the performance slightly decreased as compared with the full approach but it is still substantially ahead of Cartographer, which indicates the improvement of our approach is mainly attributed to the joint NLLS optimization of both robot poses and occupancy map simultaneously. The role of the continuous map representation as part of our approach is to establish a more accurate link between the observations and the joint NLLS optimization problem, thus can further improve the accuracy of the results slightly. While, the excellent performance of our method is mainly due to the novel joint NLLS optimization. 

\subsection{Comparisons with Cartographer using Practical Datasets}
\begin{figure*}
\centering
\includegraphics[width=1\textwidth]{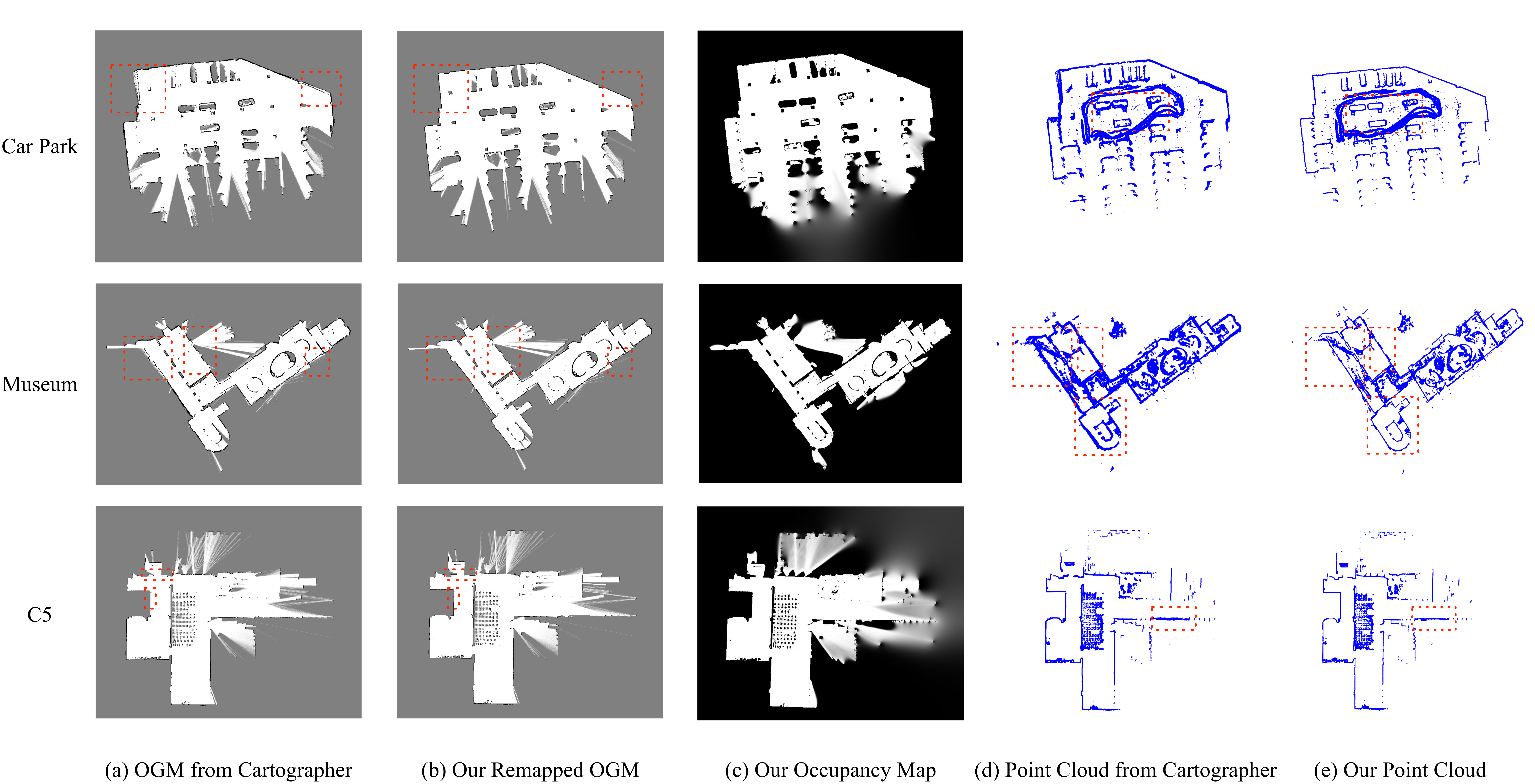}
\caption{\label{fig_result_compare} The OGMs and point cloud maps from our method and Cartographer. (a),(b),(c) show the comparison of the OGM results with Cartographer and our method. In our results, (b) shows the OGMs that we remapped using the optimized poses, and (c) shows the occupancy maps from our optimization which include some degree of smoothing. (d),(e) show the comparison of the point cloud results with Cartographer and our method. All point cloud maps are the result of projection of the scan endpoints using the optimized poses. The red dotted area shows the results of our method outperforming the results of Cartographer.}
\end{figure*}

We also use three practical datasets to compare our algorithm with Cartographer in terms of the constructed OGMs and optimized poses. Deutsches Museum dataset \cite{hess2016real} is one of the Cartographer demo datasets which includes 4144 scans and without odometry information. Car Park dataset \cite{zhao20212d} includes 1640 scans and 1639 odometry inputs collected from an underground carpark. C5 dataset is collected from a factory environment including 991 scans and 990 odometry inputs. 

We evaluated the mapping quality by comparing the details of the constructed maps. In addition, the point cloud maps constructed using the endpoint projections of the scan points of the optimized poses can be used as a reference for the accuracy of the poses to some extent.

x

The OGMs and point cloud maps of our algorithm and Cartographer using the three datasets are depicted in Fig. \ref{fig_result_compare}. 
Fig. \ref{fig_result_compare}(a) and Fig. \ref{fig_result_compare}(d) are the OGMs and point cloud maps generated from Cartographer, respectively. Fig. \ref{fig_result_compare}(b), Fig. \ref{fig_result_compare}(c) and Fig. \ref{fig_result_compare}(e) are the results from our approach, in which Fig. \ref{fig_result_compare}(c) is the occupancy map generated directly in the optimization process and Fig. \ref{fig_result_compare}(b) is the remapped OGMs using the optimized poses. Due to the smoothing term used in our approach, some blurred areas are inevitable in Fig. \ref{fig_result_compare}(c), but these parts can be completely eliminated when remapping using our optimized poses. The uncertainty maps obtained by our method are given in Fig. \ref{fig_true_uncertainty}. 
In these experiments, our method uses poses from Cartographer's results for initialization and a significant decrease in the error of the objective function was observed during the optimization process.

\begin{figure*}
\centering
\includegraphics[width=1\textwidth]{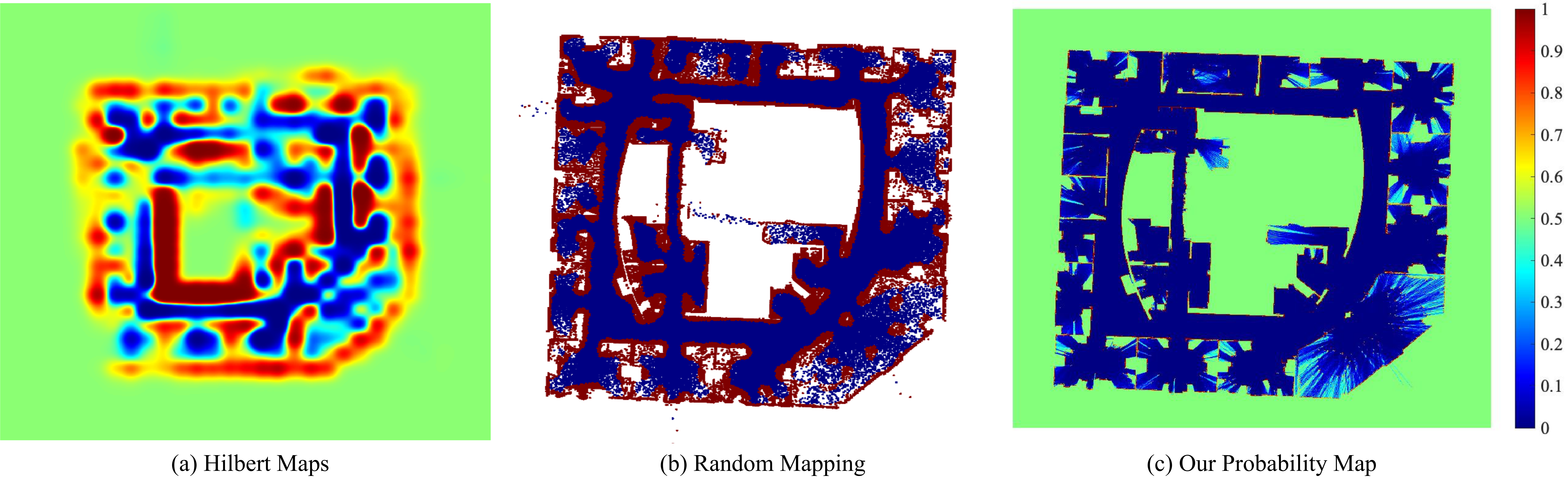}
\caption{\label{fig_probability} Comparison of our result with  Random Mapping \cite{liu2021efficient} and Hilbert Maps \cite{ramos2016hilbert} using Intel dataset.}
\end{figure*}

We use red dotted lines to highlight some areas to show our results are better than Cartographer not only in the OGMs but also in the poses. Comparing Fig. \ref{fig_result_compare}(a) with Fig.  \ref{fig_result_compare}(b), it is clear that our algorithm can obtain clearer boundaries of OGMs which thanks to our optimization of the robot poses and occupancy map simultaneously. The comparison of Fig. \ref{fig_result_compare}(d) and Fig. \ref{fig_result_compare}(e) illustrates that the poses of our method are more accurate. Although Cartographer supports loop closure detection, it still exists non-negligible errors of poses, which result in incomplete overlap of point clouds generated from observations of the same obstacle at different poses. The point cloud maps generated by our method still have incomplete overlapping point clouds, but the non-overlapping part is significantly less than that from Cartographer.      
By these experiments, it is clear that our method can further reduce the poses errors and build more accurate OGMs than Cartographer by optimizing the robot poses and occupancy map together.

\subsection{Comparison with Other Mapping Approaches using Intel Dataset}

Intel Research Lab dataset \cite{Radish} is a popular dataset used in many research papers. In this subsection, we visually compare the results from our method with two modern mapping approaches Hilbert Maps \cite{ramos2016hilbert} and Random Mapping \cite{liu2021efficient} which have both used this dataset. Since these two methods only focused on mapping with already estimated poses, aimed for quick learning process to perform the binary classification of the maps, and used different information from our method, the quantitative comparison with our method is not possible and the qualitative comparison is probably not completely fair. 

Nevertheless, we use the publicly available codes of Hilbert Maps\footnote{\url{https://bitbucket.org/LionelOtt/hilbert_maps_rss2015/src/master/} } and Random Mapping\footnote{\url{https://github.com/LiuXuSIA/RamdomMappingMethod}} with the poses given in their codes and their default parameter settings to generate the maps. The results are given in Fig. \ref{fig_probability}. Since Random Mapping method builds the map by some sampling points, it is a sparse representation and does not represent the unknown regions with a probability value of 0.5.

For better comparison, in this experiment we use probability map to present our map, which is the same map representation as Hilbert Maps \cite{ramos2016hilbert} and Random Mapping \cite{liu2021efficient}. The result is shown in Fig. \ref{fig_probability}. It is clear that our map is better than those obtained by these two modern mapping approaches for the Intel dataset. The uncertainty map obtained by our method is given in Fig. \ref{fig_true_uncertainty}. For Intel dataset, we used the robot poses from GMapping for initialization since we are not able to tune the parameters in Cartographer to obtain reasonable results\footnote{This seems to be a common issue for many researchers. See Github link \url{https://github.com/cartographer-project/cartographer_ros/issues/1420}.}.

\subsection{Robustness to Initial Guess}

\begin{figure*}
\centering
\includegraphics[width=1\textwidth]{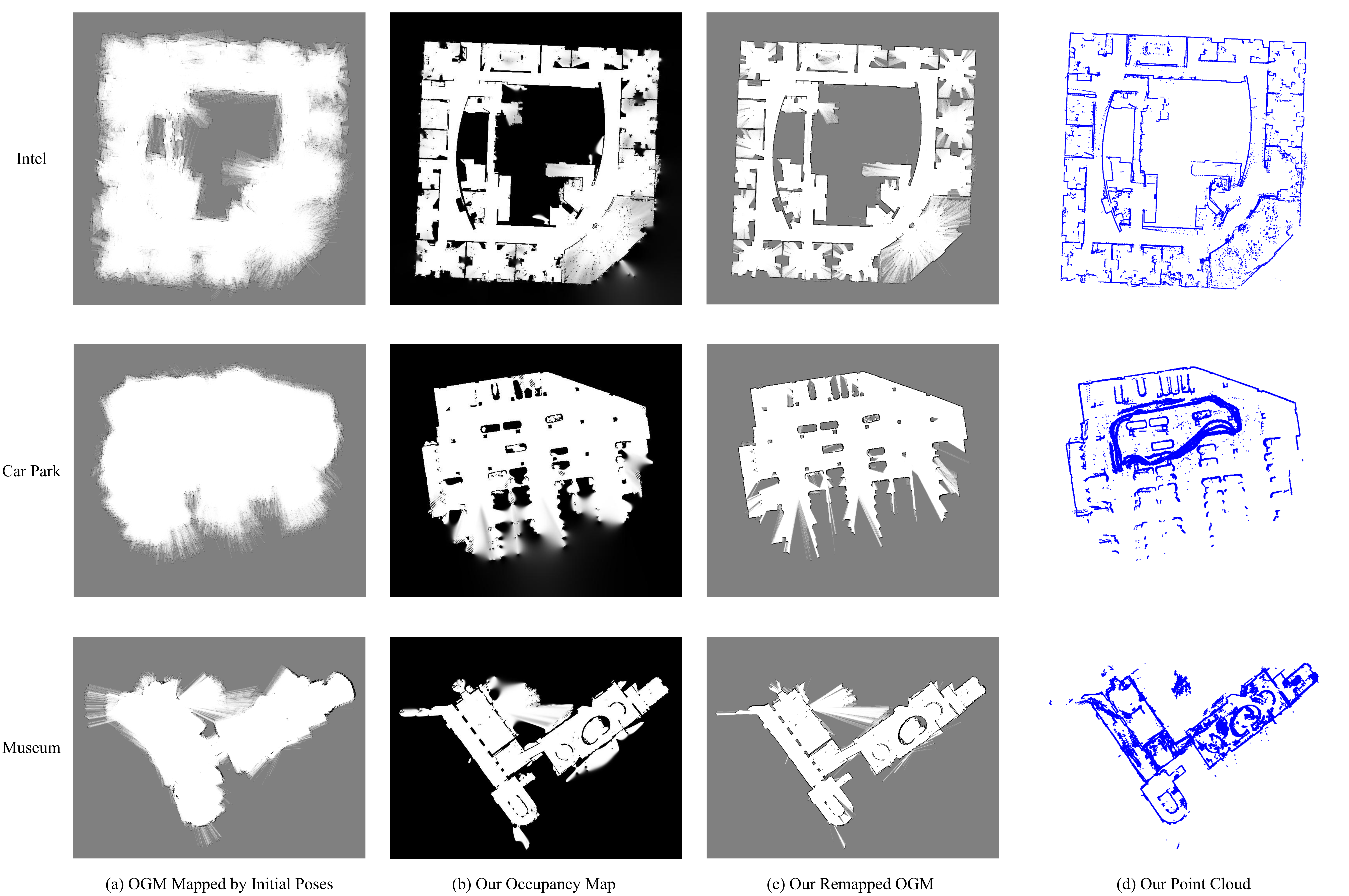}
\caption{\label{fig_bad_initial} The OGMs and point cloud maps generated from our approach using noisy poses for initialization. (a) the occupancy maps generated by the noisy poses, (b) the OGMs from our optimization, (c) the remapped OGMs built by our optimized poses, (d) the point cloud maps which projected endpoints of scans by optimized poses.}
\end{figure*}

To evaluate the robustness of our method to the initial robot poses, we use significantly noisy poses for initialization to perform robustness assessments. For Deutsches Museum, Car Park and Intel datasets, we add random noises of translation and rotation with zero-mean and standard deviation of $2 m$ for x-y and $0.5 rad$ for $\theta$ based on the poses obtained from Cartographer (for Museum and Car Park) and GMapping (for Intel). The initial occupancy maps obtained by using the noisy initial poses are shown in Fig. \ref{fig_bad_initial} (a). The convergence process of the proposed Occupancy-SLAM algorithm is shown in the supplementary video \url{https://youtu.be/4oLyVEUC4iY}.

Fig. \ref{fig_bad_initial}(b) and Fig. \ref{fig_bad_initial}(c) show the optimized occupancy maps using our method and the remapped OGMs using our optimized poses, respectively. Fig. \ref{fig_bad_initial}(d) shows the point cloud maps using our optimized poses. In this experiment, it shows our approach can converge from initial guess with large errors and also generate good results. However, starting from initial guess with significantly large errors is not guaranteed to converge to good results all the time, and sometimes the obtained “good looking” results are slightly different from those obtained from good initial values. These require further investigation.


\section{Conclusion} 
\label{sec:conclusion}

This paper formulates a new optimization based SLAM problem where the robot trajectory and the occupancy map can be optimized in one go using the information from laser scans and odometry. The optimization problem can be solved using a variation of Gauss-Newton method. Results using simulated and real experimental datasets demonstrate that our proposed method can achieve more accurate robot poses and map estimates as compared with the state-of-the-art methods. 

This paper is the first attempt to simultaneously optimize the robot poses and the occupancy map. The current approach uses uniform grid cell nodes for the map parametrization and can only serve as an offline method to achieve a high quality occupancy map. There is still a lot of work to be done to further improve the problem formulation and the solution techniques such that more robust and efficient algorithms can be achieved. Extending the work to 3D is also very interesting. These are all left to our future research work.



\bibliographystyle{plainnat}
\bibliography{references}



\end{document}